\documentclass{article}
\usepackage{ijcai19}

\usepackage{times}
\usepackage{soul}
\usepackage{url}
\usepackage[hidelinks]{hyperref}
\usepackage[utf8]{inputenc}
\usepackage[small]{caption}
\usepackage{graphicx}
\usepackage{amsmath}
\usepackage{booktabs}
\usepackage{algorithm}
\usepackage{algorithmic}
\usepackage{pdfpages}
\usepackage{color}
\usepackage{amsfonts}
\usepackage{bm}

\newtheorem{theorem}{Theorem}[section]

\newtheorem{proposition}{Proposition}[section]

\newtheorem{assumption}{Assumption}[section]

\newcommand{\qed}{\nobreak \ifvmode \relax \else
      \ifdim\lastskip<1.5em \hskip-\lastskip
      \hskip1.5em plus0em minus0.5em \fi \nobreak
      \vrule height0.75em width0.5em depth0.25em\fi}

\newcommand{\Actions}{\mathcal{A}}
\newcommand{\States}{\mathcal{S}}
\newcommand{\bw}{\mathbf{w}}
\newcommand{\bE}{\mathbb{E}}
\newcommand{\bR}{\mathbb{R}}
\newcommand{\bx}{\mathbf{x}}
\newcommand{\bd}{\hat\bx}
\newcommand{\bA}{\mathbf{A}}
\newcommand{\bc}{\mathbf{c}}
\newcommand{\bC}{\mathbf{C}}
\newcommand{\bV}{\mathbf{V}}
\newcommand{\bF}{\mathbf{F}}
\newcommand{\bb}{\mathbf{b}}
\newcommand{\bI}{\mathbf{I}}
\DeclareBoldMathCommand \bphi {\bm{\phi}}

\title{Planning with Expectation Models}

\author{
Yi Wan \footnote{denotes joint first authorship}
\and
Zaheer Abbas $^*$ \and
Adam White \and
Martha White\And
Richard S. Sutton
\affiliations
Reinforcement Learning and Artificial Intelligence Laboratory, University of Alberta\\
\emails
\{wan6, mzaheer, amw8, whitem, rsutton\}@ualberta.ca
}
\begin{document}

\maketitle

\begin{abstract}
% Two key challenges in Model-based reinforcement learning (MBRL) are 1) learning an model that captures complex stochastic dynamics of the environment 2) making reasonable plans using the learned yet inaccurate model. These two challenges, In this paper, we provide our investigate expectation model and a related planning algorithm for. In particular, we: 1) show that planning with an expectation model is equivalent to planning with a distribution model if the state value function is linear in state features 2) study two common parametrization choices for approximating the expectation: linear and non-linear expectation models 3) propose a model-based policy evaluation algorithm and present its convergence results 4) empirically demonstrate the soundness of the proposed planning algorithm.
Distribution and sample models are two popular model choices in model-based reinforcement learning (MBRL). However, learning these models can be intractable, particularly when the state and action spaces are large. Expectation models, on the other hand, are relatively easier to learn due to their compactness and have also been widely used for deterministic environments. For stochastic environments, it is not obvious how expectation models can be used for planning as they only partially characterize a distribution. In this paper, we propose a sound way of using expectation models for MBRL. In particular, we 1) show that planning with an expectation model is equivalent to planning with a distribution model if the state value function is linear in state-feature vector, 2) analyze two common parametrization choices for approximating the expectation: linear and non-linear expectation models, 3) propose a sound model-based policy evaluation algorithm and present its convergence results, and 4) empirically demonstrate the effectiveness of the proposed planning algorithm. % \footnote{Supplementary materials: https://arxiv.org/abs/1904.01191} 
   % we show that when value function is linear, though convergent point of planning with both linear and non-linear expectation model is equivalent to it with the true model for on-policy policy evaluation, only non-linear expectation model is equivalent to true model for off-policy policy evaluation
\end{abstract}

\section{Introduction}

% In this paper we focus on two unsolved problems in MBRL.
% \begin{itemize}
%     \item what kind of model is preferred and how to learn it.
%     \item when model is approximated, how to deal with the distribution mismatch problem between model training and applying
% \end{itemize}

% There are two related challenges in MBRL: how to learn a model for a generally stochastic environment and how to use a learned potentially flawed model for planning in a sound way.

% We propose a set of convergent planning algorithms for policy evaluation which involves expectation model + linear value functions and is based on a new objective function: Model based MSPBE.

Learning models of the world and effectively planning with them remains a long-standing challenge in artificial intelligence. Model-based reinforcement learning formalizes this problem in the context of reinforcement learning where the \textit{model} approximates the environment's dynamics. %Once the model of the environment has been learned, an agent can use it for planning.
The output of the model is one of the key choices in the design of a planning agent, as it determines the way the model is used for planning. Should the model produce 1) a distribution over the next-state feature vectors and rewards, 2) a sample of the next-state feature vector and reward , or 3) the expected next-state feature vector and reward? For stochastic environments, distribution and sample models can be used effectively, particularly if the distribution can be assumed to be of a special form \cite{deisenroth2011pilco,chua2018deep}. For arbitrarily stochastic environments, learning a sample or distribution model could be intractable or even impossible. For deterministic environments, expectation models appear to be the default choice as they are easier to learn and have been used \cite{oh2015action,leibfried2016deep}. However, for general stochastic environments, it is not obvious how expectation models can be used for planning as they only partially characterize a distribution. %TODO: can we cite something for this?
In this paper, we develop an approach to use expectation models for arbitrarily stochastic environments by restricting the value function to be linear in the state-feature vector.

Once the choice of expectation models with linear value function has been made, the next question is to develop an algorithm which uses the model for planning. In previous work, planning methods have been proposed which use expectation models for policy evaluation \cite{sutton2012dyna}.  However, as we demonstrate empirically, the proposed methods require strong conditions on the model which might not hold in practice, causing the value function to diverge to infinity. Thus, a key challenge is to devise a \textit{sound} planning algorithm which uses expectation models for policy evaluation and has convergence guarantees. In this work, we propose a new objective function called \textit{Model Based-Mean Square Projected Bellman Error} (MB-MSPBE) for policy evaluation and show how it relates to \textit{Mean Square Projected Bellman Error} (MSPBE) \cite{sutton2009fast}. We derive a planning algorithm which minimizes the proposed objective and show its convergence under conditions which are milder than the ones assumed in the previous work \cite{sutton2012dyna}. 

It is important to note that in this work, we focus on the value prediction task for model-based reinforcement learning. Predicting the value of a policy is an integral component of \textit{Generalized Policy Iteration} (GPI) on which much of modern reinforcement learning control algorithms are built \cite{sutton2018reinforcement}. Policy evaluation is also key for building predictive world knowledge where the questions about the world are formulated using value functions \cite{sutton2011horde,modayil2014multi,white2015developing}. More recently, policy evaluation has also been shown to be useful for representation learning where value functions are used as auxiliary tasks \cite{jaderberg2016reinforcement}. While model-based reinforcement learning for policy evaluation is interesting in its own right, the ideas developed in this paper can also be extended to the control setting.% where the learned models can be used for greedy action-selection leading to a \textit{GPI-like} algorithm \cite{sutton2012dyna}.

% The main contributions of this paper are as follows: 1) we show that planning with an expectation model is equivalent to planning with a distribution model if the state value function is linear in state features, 2) we study two common parametrization choices for approximating the expectation: linear and non-linear expectation models, 3) we propose a model-based policy evaluation algorithm and present its convergence results, and  4) we empirically demonstrate the soundness of the proposed planning algorithm.

% for on-policy policy evaluation TD(0) planning (or for off-policy policy evaluation GTD-Model) with the \text{best} linear expectation model converges to the same solution as the best non-linear model if the state-value function is linear 
% An approximate model takes the state's feature vector and action as input and predicts how the environment would respond to the agent's action. The output of the model is one of the key choices in the design of a planning agent which determines the way the model is used for planning. Should the model produce the distribution of the resulting-state feature vectors, a sample of the feature vector, or the expected value of the feature vector? 

\section{Problem Setting}

We formalize an agent's interaction with its environment by a finite Markov Decision Process (MDP) defined by the tuple $(\mathcal{S}, \mathcal{A}, \mathcal{R}, p, \gamma)$, where $\mathcal{S}$ is a set of states, $\mathcal{A}$ is a set of actions, $\mathcal{R}$ is a set of rewards, $\gamma \in [0,1)$ is a discount factor, and $p: \mathcal{S} \times \mathcal{A} \times \mathcal{S} \times \mathcal{R} \mapsto [0, 1]$ is the dynamics of the environment such that $p(s', r| s, a) \doteq \Pr (S_{t+1} = s', R_{t+1} = r | S_t = s, A_t = a)$ for all $s, s' \in \mathcal{S}, a \in \mathcal{A}$, and $r \in \mathcal{R}$. A stationary policy $\pi$ determines the behavior of the agent. The value function $v_\pi:\mathcal{S}\mapsto\mathbb{R}$ describes the expected discounted sum of rewards obtained by following policy $\pi$ from each state.

In function approximation setting, value, policy and model are all functions of state, but only through a $m$-dimensional real-valued feature vector $\bx_t = \bx(S_t)$ where $\bx: \mathcal{S} \mapsto \mathbb{R}^m$ is the feature mapping, which can be an arbitarily complex function.
Tile-coding \cite{sutton1996generalization} and Fourier basis \cite{konidaris2011value} are examples of state-feature mapping functions which are expert designed. An alternative is to learn the mapping $\bx$ using auxiliary tasks \cite{chung2018two,jaderberg2016reinforcement}. The policy $\pi : \bR^m \times \Actions \to [0, 1]$ is usually a parameterized function mapping from a $m$-dimensional feature vector and an action to the probability of choosing the action when observing the feature vector. The value function is usually approximated using a parametrized function with a $n$-dimensional weight vector $\bw\in \mathbb{R}^n$, where typically $n \ll |\mathcal{S}|$. We write $\hat{v}(\bphi, \bw) \approx v_\pi(s)$ for the approximate value of state $s$, if $\bphi$ is the feature vector of $s$. The approximate value function can either be a linear function of the state-feature vector $\hat{v}(\bphi, \bw) = \bphi^\top \bw$ where $n = m$, or a non-linear function $\hat{v}(\bphi, \bw) = f(\bphi, \bw)$ where $f: \mathbb{R}^m \times \mathbb{R}^n \mapsto \mathbb{R}$ is an arbitrary function. In addition, the state feature vector is also used as both input and output of the model, which we will discuss in the next section. % Deep Q-Network \cite{mnih2015human} for playing ALE \cite{bellemare2013arcade} is an example of non-linear value function approximation where the last four stacked frames serve as an expert designed state-feature representation.

The Dyna architecture \cite{sutton1991integrated} is an MBRL algorithm which unifies learning, planning, and acting via updates to the value and policy functions. The agent interacts with the world, using observed state, action, next state, and reward tuples to learn a model $\hat p$ to capture the environment dynamics $p$, and update value and policy functions. The planning step in Dyna repeatedly produces predicted next states and rewards from the model, given input state-action pairs. These hypothetical experiences can be used to update the value and policy functions, just as if they had been generated by interacting with the environment. The {\em search control} process decides what states and actions are used to query the model during planning. The efficiency of planning can be significantly improved with clever search control strategy such as prioritized sweeping \cite{moore1993prioritized,sutton2012dyna,pan2018organizing}. In the function approximation setting, there are three factors that can affect the solution of a planning algorithm: 1) the distribution of data used to train the model, 2) the search control process's distribution for selecting the starting feature vectors and actions for simulating the next feature vectors, and 3) the policy being evaluated.

Consider an agent wanting to evaluate a policy $\pi$, i.e., approximate $v_\pi$, using a Dyna-style planning algorithm. Assume that the data used to learn the model come from the agent's interaction with the environment using policy $b$. It is common to have an ergodicity assumption on the markov chain induced by $b$:
\begin{assumption} \label{behavior policy assumption}
The markov chain induced by policy b is ergodic.
\end{assumption}

Under this assumption we can define the expectation $\bE_b[\cdot]$ in terms of the unique stationary distribution of $b$, for example, $\bE_b[\bx_t \bx_t^\top] = \sum_{s \in \States} \eta(s) \bx(s) \bx(s)^\top$ where $\eta$ denote $b$'s stationary state distribution. % It is implicitly assumed whenever we write $\bE_b$ in this paper. 

Let $H_{\bphi} \doteq \{s \in \States: \bx(s) = \bphi \}$ as the set of all states sharing feature vector $\bphi$. Consequently, the stationary feature vector distribution corresponding to $\eta$ would be $\mu(\bphi) \doteq \sum_{s \in H_{\bphi}} \eta(s)$. Let's suppose the search control process generates a sequence of i.i.d random vectors $\{\bphi_k \}$ where each $\bphi_k$ is a $m$-dimensional random vector following distribution $\zeta$, and chooses actions $\{A_k\}$ according to policy $\pi$ i.e. $A_k \sim \pi(\cdot | \bphi_k)$, which is the policy to be evaluated. The sequence $\{\bphi_k\}$ is usually assumed to be bounded.

\begin{assumption} \label{bounded feature vector}
$\{\bphi_k\}$ is bounded.
\end{assumption}

For the uniqueness of the solution, it is also assumed that the features generated by the search-control process are linearly independent:

\begin{assumption} \label{search control assumption}
$\bE_\zeta[\bphi_k \bphi_k^\top] $ is non-singular.
\end{assumption}

\section{Models}
A model can be learned if the environment dynamics is not known. There are at least three types of models: distribution models, sample models and expectation models. While all of them take a state-feature vector $\bphi$ and an action $a$ as input, their outputs are different. 

A distribution model $\hat p$ produces the distribution over the next state-feature vectors and rewards: $\hat p(\bphi', r |\bphi,a) \approx {\bf Pr}[\bx_{t+1}=\bphi', R_{t+1} = r|\bx_t\!=\!\bphi, A_t\!=\!a]$.
Distribution models have typically been used with special forms such a Gaussians \cite{chua2018deep} or Gaussian processes \cite{deisenroth2011pilco}. In general, however, learning a distribution can be impractical as distributions are potentially large objects. For example, if the state is represented by a feature vector of dimension $m$, then the first moment of its distribution is a $m$-vector, but the second moment is a $m \times m$ matrix, and the third moment is $m \times m \times m$, and so on. 
%Unless the distribution can be assumed to be of known special form, these are very large objects. In general, all the moments are required to characterize the distribution, but that would be hopelessly unwieldy for large $m$. 

Sample models are a more practical alternative, as they only need to generate a sample of the next-state feature vector and reward, given a state-feature vector and action. 
%generally do not need to represent the full distribution, but 
Sample models can use arbitrary distributions to generate the samples---even though they do not explicitly represent those distributions---but can still produce objects of a limited size (e.g. feature vectors of dimension $m$).  
%An alternative that is more practical in some ways is for the model to return not the full distribution, but a sample from it. Then a sense of the full distribution can be obtained incrementally by repeated sampling. 
They are particularly well suited for sample-based planning methods such as Monte Carlo Tree Search \cite{coulom2006efficient}. 
% MARTHAC: Is this true?
%Sampling models can use arbitrary distributions to generate the samples but can still produce objects of a limited size (e.g. feature vectors of dimension $m$). In addition, it is easier to iterate sample models, that is, to take the state-feature output of the model's projection and feed it in again as the state-feature input, projecting to a sample state-feature of two iterations later, and so on. 
Unlike distribution models, however, sample models are stochastic which creates an additional branching factor in planning, as multiple samples are needed to be drawn to gain a representative sense of what might happen. 

% Generative Adversarial Networks \cite{goodfellow2014generative} and Variational Autoencoders \cite{kingma2013auto} are two powerful frameworks to learn sample models for complex empirical distributions. While they have shown encouraging results for model-based reinforcement learning \cite{corneil2018efficient}, they are still restricted by a neural network's capacity and trainability in the extent to which they can approximate arbitrarily complex high-dimensional distributions. In addition, they are known to struggle with discrete distributions since backpropagation through discrete variables is generally not possible. While improving on these methods is a promising research direction and a topic of great interest with exciting ideas, we now turn towards expectation models, which offers an attractive avenue for planning, and discuss its implications for model-based reinforcement learning.
% TODO: cite GANs/VAEs papers for discrete distributions/improving on trainability/avoiding mode collapse

% We should note that methods exist which use a distributional model of a know special form and then draw samples from the distribution for planning \cite{deisenroth2011pilco}\cite{chua2018deep}.

%\section{Expectation Model}
Expectation models are an even simpler approach, where the model produces the \emph{expectation} of the next reward and the next-state
feature vector. An expectation model $\{\hat \bx, \hat r\}$ consists of two functions: feature-vector function $\hat\bx$ such that $\hat\bx(\bphi,a)\approx\mathbb{E}_b[\bx_{t+1}|\bx_t\!=\!\bphi, A_t\!=\!a]$ and reward function $\hat r$ such that $\hat r(\bphi, a) \approx \mathbb{E}_b[R_{t+1}|\bx_t\!=\!\bphi, A_t\!=\!a]$.
%We call this kind of model an \emph{expectation model}. 
The advantages of
expectation models are that the feature vector output is compact (like a sample model) and
deterministic (like a distribution model). The potential disadvantage of an expectation model is that it is only a partial characterization of the distribution. For example, if the result of an action (or option) is that two binary state features both occur with probability 0.5, but are never present (=1) together, then an expectation model can capture the first part (each present
with probability 0.5), but not the second (never both present). This may not be a
substantive limitation, however, as we can always add a third binary state feature,
for example, for the AND of the original two features, and then capture the full
distribution with the expectation of all three state features.

% MARTHAC: maybe we should leave it with the example of augmenting the state? We don't say more about expressivity, but here we suggest we are going to explore it
%The expectation model puts more pressure on the mapping $\bx$ for extracting state-features, but it is also a substantial relaxation of what is required from a distribution and a sample model. Both of these other models have to find and represent any correlations or anti-correlations between the state features. Overall it is not clear that the amount of work required changes when moving from a distribution or sampling model to an expectation model; more likely the burden is just shifted from constructing the sample/distribution model to constructing the state representation. In general, the state representation (feature vector) can be expected to be larger (to have more components) with an expectation model.

\section{Expectation Models and Linearity}
Expectation models can be less complex than distribution and sample models and, therefore, can be easier to learn. This is especially critical for model-based reinforcement learning where the agent is to learn a model of the world and use it for planning. In this work, we focus on answering the question: how can expectation models be used for planning in Dyna, despite the fact that they are only a partial characterization of the distribution?
%Since expectation models are only a partial characterization of the transition dynamics, a natural question is to ask how they can be used for planning.

There is a surprisingly simple answer to this question: if the value function is linear in the feature vector, then there is no loss of generality when using an expectation model instead of a distribution model for planning. 
Let $\hat p$ be an arbitrary distribution model and $\hat\bx, \hat r$ be the feature-vector and reward functions of the corresponding expectation model, that is, with $\hat\bx(\bphi,a)=\sum_{\bphi', r}\bphi'\hat p(\bphi', r|\bphi,a)$ and $\hat r(\bphi,a)=\sum_{\bphi', r} r \hat p(\bphi', r|\bphi,a)$. % Let both models share the same $\hat r$.
%It is easy to show this formally, starting from planning by value iteration. Plugging the linear value function into the general distribution-model form of value iteration yields
Then approximate dynamic programming (policy evaluation) with the distribution model and linear function approximation performs an update of $\bw$ for each feature vector $\bphi$ such that $\hat{v}(\bphi, \bw)$ is moved toward:
\begin{align}
&\sum_{a}\pi(a|\bphi) \sum_{\bphi', r} \hat{p}(\bphi', r|\bphi, a) \bigl[r + \gamma  \hat{v}(\bphi', \bw)\bigr] \label{value iteration with distribution model}\\
&=\sum_{a}\pi(a|\bphi) \biggl[\hat r(\bphi,a) + \gamma\sum_{\bphi', r} \hat{p}(\bphi', r|\bphi, a) \bphi'^\top\bw\biggr]\nonumber \\
&=\sum_{a}\pi(a|\bphi) \bigl[\hat r(\bphi,a) + \gamma \hat\bx(\bphi,a)^\top\bw\bigr]\label{value iteration with expectation model}
\end{align}
The last expression, using just an expectation model, is equal to the first, thus showing that no generality has been lost compared to an arbitrary distribution model, if the value function is linear. Further, the same equations also advocate the other direction: if we are using an expectation model, then the approximate value function should be linear. This is because \eqref{value iteration with expectation model} is unlikely to equal \eqref{value iteration with distribution model} for general distributions if $\hat{v}$ is not linear in state-feature vector.

It is important to point out that linearity does not particularly restrict the expressiveness of the value function since the mapping $\bx$ could still be non-linear and, potentially, learned end-to-end using auxiliary tasks \cite{jaderberg2016reinforcement,chung2018two}.  % Finally, note that similar derivation can be done for approximating $v_\pi$ using value iteration.
% TODO: Transition which summarizes what is to come next

\section{Linear \& Non-Linear Expectation Models}
We now consider the parameterization of the expectation model: should the model be a linear projection from a state feature vector and an action to the expected next state feature vector and reward or should it be an arbitrary non-linear function? In this section, we discuss the two common choices and their implications in detail. 

The general case for an expectation model is that $\hat\bx$ and $\hat r$ are arbitrary non-linear functions. A special case is of the linear expectation model, in which both of these functions are linear, i.e., $\hat\bx(\bphi, a) = \bF_a \bphi$ and $\hat r(\bphi, a) = \bb_a^\top \bphi$ where matrices $\{\bF_a, \forall a \in \Actions\}$ (shortened as $\{\bF_a\}$) and vectors $\{\bb_a, \forall a \in \Actions\}$ (shortened as $\{\bb_a\}$) are parameters of the model.

%We later present empirical results showing that planning with linear expectation models converges faster when compared with the non-linear model, when both models are learned online. Though not extensive, our results demonstrate that linear model and linear value function could be an attractive design choice for model-based reinforcement, especially when coupled with a non-linear  mapping $\bx$ for state-features. In these senses, the choices of an expectation model, a linear approximate value function, and a linear model are tightly linked. It is not a rigorous argument that one must make any one of these choices on its own, but once some have been made, there are strong reasons to adopt all of them. Let us call planning based on all three choices \textit{expectation-based planning}.

%\subsection{Theory for on-policy policy evaluation}

We now define the best linear expectation model and the best non-linear expectation model trained using data generated by policy $b$. In particular, let $\{\bF_a^*\}, \{\bb_a^*\}$ be the parameters of best linear expectation model. Then we have $\forall a \in \Actions$

\begin{align*}
    \bF_a^* & \doteq \arg \min_{\mathbf{G}} \bE_b [\hspace{0.05cm} \mathbb{I}(A_t = a) \Vert \mathbf{G} \bx_t - \bx_{t+1} \Vert_2^2 \hspace{0.05cm}]\\
    \bb_a^* & \doteq \arg \min_{\mathbf{u}} \bE_b [\hspace{0.05cm} \mathbb{I}(A_t = a) (\mathbf{u}^\top \bx_t - R_{t+1} )^2 \hspace{0.05cm}] 
\end{align*}

For the uniqueness of the best linear model, we assume
\begin{assumption} \label{model learning assumption}
$\bE_b[\mathbb{I}(A_t = a)\bx_t \bx_t^\top]$ is non-singular $\forall a \in \Actions$
\end{assumption}
Under this assumption we have closed-from expression of $\bF_a^*, \bb_a^*$
\begin{align*}
    \bF_a^* & = \bE_b [\mathbb{I}(A_t = a) \bx_{t+1} \bx_{t}^\top] \bE_b [\mathbb{I}(A_t = a)  \bx_t \bx_t^\top]^{-1}\\
    \bb_a^* & = \bE_b [\mathbb{I}(A_t = a)  \bx_t \bx_t^\top]^{-1} \bE_b [\mathbb{I}(A_t = a) \bx_t  R_{t+1}]
\end{align*}

Let $\hat \bx^*, \hat r^*$ be the feature-vector and reward functions of the best non-linear expectation model. Then we have
\begin{align*}
    \hat\bx^* (\bphi, a) & \doteq \bE_b [\bx_{t+1}| \bx_t = \bphi, A_t = a] \nonumber \\
    & = \frac{\sum_{s \in H_{\bphi}} \eta(s) \bE_b [\bx(S_{t+1}) | S_t = s, A_t = a]}{\mu(\bphi) }\\
    \hat r^*(\bphi, a) & \doteq \bE_b [R_{t+1} | \bx_t = \bphi, A_t = a] \nonumber \\
    & = \frac{\sum_{s \in H_{\bphi}} \eta(s) \bE_b [R_{t+1} | S_t = s, A_t = a]}{\mu(\bphi) }
\end{align*}

Both linear and non-linear models can be learned using samples via stochastic gradient descent.

%We present our first theorem showing on-policy TD(0) planning with both the best linear and non-linear expectation model converge to the TD-fixed point under certain conditions.
%
%\begin{theorem}[Convergence of on-policy TD(0) planning]\label{Convergence of On-policy TD(0) Planning with the Expectation Model}
%Given a stationary policy $b \in \mathbb{R}^m \mapsto [0, 1]^{|\mathcal{A}|}$, assume:
%\begin{enumerate}
%    \item The markov chain induced by policy $b$ is ergodic with stationary state distribution $\eta$  and corresponding feature vector distribution $\mu$
%    \item $(\bphi_k)$ are i.i.d random variables and  $\bphi_k \sim \mu$.
%    \item $(A_k)$ are i.i.d random variables and  $A_k \sim b(\cdot | \bphi_k)$.
%    \item $\bE_b[\bx_t \bx_t^\top] = \bE [\bphi_k \bphi_k^\top]$ is non-singular.
%    \item $\alpha_k > 0$, $\sum_{k=0}^\infty \alpha_k = \infty$ and $\sum_{k=0}^\infty \alpha_k^2 < \infty$.
%    \item $\hat\bx(\bphi, a) = \bF_a^* \bphi$ or $\hat\bx(\bphi, a) = \hat\bx^*(\bphi, a) $
%    \item $r(\bphi, a) = \bb_a^* \bphi$  or $r(\bphi, a) = r^*(\bphi, a) $.
%\end{enumerate}
%Then iteration:
%\begin{align*}
%    \bw_{k+1} \leftarrow \bw_{k} + \alpha_k (r(\bphi_k, A_k) + \gamma \bw_k^\top \hat\bx(\bphi_k, A_k)  - \bw_k^\top \bphi_k) \bphi_k
%\end{align*}
%converges to the TD-fixed point.
%\end{theorem}

%\subsection{Empirical results for on-policy policy evaluation}
% \subsubsection{Utility of Planning}
% \textcolor{red}{Empirical results for showing the utility of planning (sample-efficiency) go here.}

\subsection{Why Linear Models are Not Enough?}
In previous work, linear expectation models have been used to simulate a transition and execute TD(0) update \cite{sutton2012dyna}. Convergence to the TD-fixed point using TD(0) updates with a \textit{non-action linear expectation model} is shown in theorem 3.1 and 3.3 of \cite{sutton2012dyna}.
An additional benefit of this method is that the point of convergence does not rely on the distribution $\zeta$ of the search-control process. Critically, a non-action model cannot be used for evaluating an arbitrary policy, as it is tied to a single policy -- the one that generates the data for learning the model. To evaluate multiple policies, an action model is required. In this case, the point of convergence of the TD(0) algorithm is dependent on $\zeta$. From corollary 5.1 of \cite{sutton2012dyna}, the convergent point of TD(0) update with linear action model with parameters $\{\bF_a\}, \{\bb_a\}$ is:
\begin{align}\label{best linear model solution}
(\bI - \gamma \bF^\top)^{-1} \bb
\end{align}
where  $\bF = \bE_\zeta[\bF_{A_k} \bphi_k \bphi_k^\top] \bE_\zeta[\bphi_k \bphi_k^\top]^{-1}, \bb = \bE_\zeta[\bphi_k \bphi_k^\top]^{-1} \bE_\zeta[ \bphi_k \bphi_k^\top \bb_{A_k}] $. It is obvious that the convergence point changes as the feature vector generating distribution $\zeta$ changes. We now ask even if $\zeta$ equals $\mu$, is the solution of planning using the model same as the solution of learning using the real data. In the next proposition we show that this is not true in general for the best linear model, however, it is true for the best non-linear model.

%  the fixed point of the expected TD(0) update using the data generated by the model same as the fixed-point of the expected off-policy TD(0) learning update using the real data, assuming both fixed points exist. 
% The next question is how good the convergence point is for TD(0) updates with the best linear model. For example, is that still the same as it with the real environment, as it does in the degenerated non-action model case? 

% If converges, TD(0) update converges to the TD-fixed point. 
Let the fixed point of the expected off-policy TD(0) update using the real data be $\bw_{\text{env}}$, the fixed point of the expected TD(0) update using the data generated by the best linear model be $\bw_{\text{linear}}$ , and the fixed point of the expected TD(0) update using the data generated by the best non-linear model be $\bw_{\text{non-linear}}$ (assuming three fixed points all exist). These fixed points are all unique and can be written as follow:
\begin{align}
    & \bw_{\text{env}} = \bE_b[\rho_t \bx_t (\bx_t - \gamma \bx_{t+1})^\top]^{-1} \bE_b[\rho_t R_{t+1} \bx_t] \label{real system solution}\\
    & \bw_{\text{non-linear}} = \bE_\zeta [\bphi_k ( \bphi_k - \gamma \bd^*(\bphi_k, A_k))^\top ]^{-1} \bE_\zeta [\hat r^*(\bphi_k, A_k) \bphi_k] \nonumber \\
    & \bw_{\text{linear}} =  \text{equation (\ref{best linear model solution}) with } \bF_a = \bF_a^*, \bb_a = \bb_a^*  \nonumber
\end{align}
where $\rho_t = \frac{\pi(A_t | \bx_t)}{b(A_t | \bx_t)}$ is the importance sampling ratio.

\begin{proposition}\label{TD(0) planning}
suppose
\begin{enumerate}
    \item assumptions \ref{behavior policy assumption}, \ref{model learning assumption} hold
    \item $\zeta = \mu$.
\end{enumerate}
then in general $\bw_{\text{env}} = \bw_{\text{non-linear}} \neq \bw_{\text{linear}}$
% then if in addition $\bd = \bd^*, r =  r^*$
% \begin{align*}
%   \bE [\delta_k \bphi_k] = \bE [\delta_t \bx_t| S_t \sim \eta(\cdot), A_t \sim \pi(\cdot|\bx_t)]
% \end{align*}
% however if $\bd(\bphi, a) = \bF_a^* \bphi, r(\bphi, a) = \bb_a^{*\top} \bphi$.
% \begin{align*}
%   \bE [\delta_k \bphi_k] \neq \bE [\delta_t \bx_t| S_t \sim \eta(\cdot), A_t \sim \pi(\cdot|\bx_t)]
% \end{align*}
% in general.
\end{proposition}

 \subsection{An Illustrative Example on the Limitation of Linear Models}
\begin{figure}\centering
  \includegraphics[width=7cm]{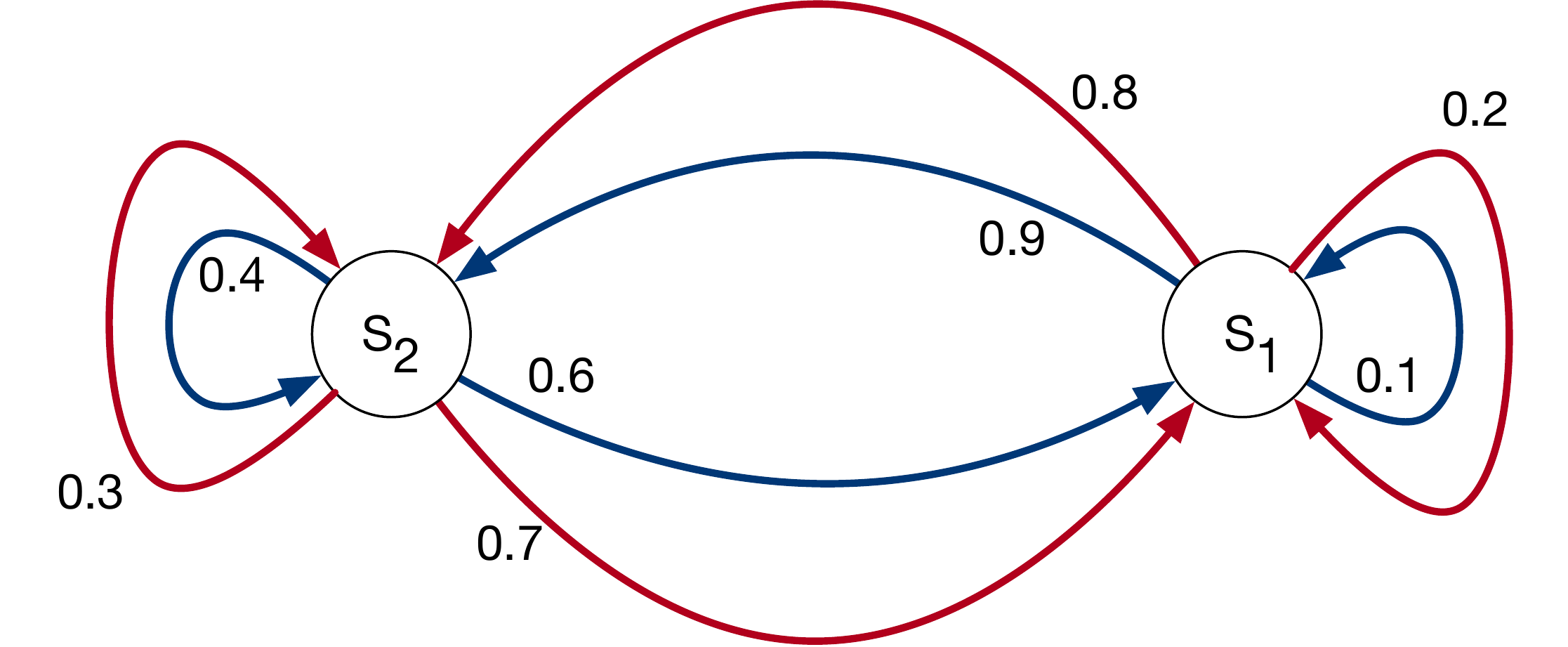}
  \caption{A simple two state MDP in which two actions, \textit{blue} and \textit{red}, are available in each state. Each action causes the agent to probabilistically move to the next-state or stay in the same state. The specific transition probabilities are specified next to the arrows representing the transitions. The reward is zero everywhere except when the \textit{red} action is taken in state $s_1$. The feature vector has only one component $\bx(s_1)=[0.5]^\top$ and $\bx(s_2)=[-0.1]^\top$ The policy $b$ used to generate the data for learning the model is given as: $b(\cdot |\bx(s_1)) = [0.1, 0.9], b(\cdot|\bx(s_2)) = [0.3, 0.7]$. The policy $\pi$ to be evaluated is given by: $\pi(\cdot|\bx(s_1)) = [0.4, 0.6], \pi(\cdot|\bx(s_2)) = [0.5, 0.5]$. }
  \label{fig:MDP}
\end{figure}

In order to clearly elucidate the limitation of linear models for planning, we use a simple two-state MDP, as outlined in Figure \ref{fig:MDP}. The policy $b$ used to generate the data for learning the model, and the the policy $\pi$ to be evaluated are also described in Figure \ref{fig:MDP}. We learn a linear model with the data collected by interacting with the environment using policy $b$ and verify that it is the best linear model that could be obtained. We can then obtain $\bw_{\text{linear}}$ using equation(\ref{best linear model solution}). $\bw_{\text{env}}$ is calculated by \textit{off-policy LSTD}\cite{yu2010convergence} using the same data that is used to learn the linear model. In agreement to proposition \ref{TD(0) planning}, the two resulting fixed points are different: $\bw_{\text{linear}} = [0.953]^\top$ and $\bw_{\text{env}} = [8.89]^\top$. 

Previous works \cite{parr2008analysis,sutton2012dyna} showed that a non-action linear expectation model could just be enough if the value function is linear in state-feature vector. Proposition \ref{TD(0) planning} coupled with the above example suggests that this is not true for the more general case of linear expectation models, and expressive non-linear models could potentially be a better choice for planning with expectation models. From now on, we focus on non-linear models as the parametrization of choice. % for planning with expectation models.

\section{Gradient-based Dyna-style Planning Methods}

In the previous section, we established that more expressive non-linear models are needed to recover the solution obtained by real data. An equally crucial choice is that of the planning algorithm: do TD(0) updates converge to the fixed-point? For this to be true for linear models, the numerical radius of $\bF$ needs to be less than 1 \cite{sutton2012dyna}. We conjecture that this condition might not hold in practice causing the planning to diverge and illustrate this point using \textit{Baird's Counter Example} \cite{baird1995residual} in the next section. 
% It might not hold either if we vary the search control strategy or just because the model is flawed.

The proof of proposition \ref{TD(0) planning} further implies that the expected TD(0) update with the best non-linear model $\bE_\zeta [\Delta_k \bphi_k]$ is the same as the expected model-free off-policy TD(0) update $\bE_b[\rho_t \delta_t \bx_t]$, where $\Delta_k = \hat r(\bphi_k, A_k)+ \gamma \bw^\top \hat\bx(\bphi_k, A_k) - \bw^\top \bphi_k$ and $\delta_t = R_{t+1} + \gamma \bw^\top \bx_{t+1} - \bw^\top \bx_t$. However, for off-policy learning, TD(0) is not guaranteed to be stable, which suggests that even with the best non-linear model, the TD(0) algorithm is also prone to divergence. This is also empirically verified in Baird’s Counter Example.

% On the other hand, from the above proposition we see the expected TD(0) update with the best non-linear model  and the real environment is the same,  where . This means even if one uses the best non-linear model, the convergence issue still exists, due to the same problem as off-policy TD(0) learning.

Inspired by the \emph{Gradient-TD} off-policy policy evaluation algorithms \cite{sutton2009fast} which are guaranteed to be stable under function approximation, we propose a family of convergent planning algorithms. The proposed methods are guaranteed to converge for both linear and non-linear expectation models. This is true even if the models are imperfect, which is usually the case in model-based reinforcement learning where the models are learned online.

To achieve this goal, we define a new objective function which our planning algorithm would optimize: \emph{Model-Based Mean Square Projected Bellman Error}, $\text{MB-MSPBE}(\bw) =  \bE_\zeta [\Delta_k \bphi_{k}]^\top \bE_\zeta[\bphi_k \bphi_k^\top]^{-1} \bE_\zeta [\Delta_k \bphi_{k}]$. This new objective is similar to what GTD learning algorithm optimizes, the Mean Square Projected Bellman Error, $\text{MSPBE}(\bw) = \bE_b [\rho_t \delta_t \bx_{t}]^\top \bE_b[\bx_t \bx_t^\top]^{-1} \bE_b [\rho_t \delta_t \bx_{t}]$. This new objective can be optimized using a variety of gradient-based methods -- as we will elaborate later. We call the family of methods using gradient to optimize this objective \emph{Gradient-based Dyna-style Planning} (Gradient Dyna) methods.

Similar to MSPBE, MB-MSPBE is also quadratic, and if it is not strictly convex then minimizing it give us infinite solutions for $\bw$. Since features are assumed to be independent, this would mean that we have infinite different solutions for the approximate value function. Therefore it's reasonable to assume that the solutions of minimizing MB-MSPBE is unique. This would be true iff the Hessian  $\nabla^2 \text{MB-MSPBE}(\bw) = \bA^{\top} \bC^{-1} \bA$, where $\bA = \bE_\zeta [\bphi_k ( \bphi_k - \gamma \hat\bx(\bphi_k, A_k))^\top ]$ and $\bC = \bE_\zeta[\bphi_k \bphi_k^\top]^{-1}$, is invertible. This is equivalent to $\bA$ being non-singular. 

\begin{assumption}\label{MB-MSPBE assumption}
$\bA$ is non-singular.
\end{assumption}
It can be shown that the unique solution for minimizing this new objective is $\bw^* \doteq \bA^{-1} \bc$, where $\bc = \bE_\zeta [\hat{r}(\bphi_k, A_k) \bphi_k]$, if the above assumption holds.

Furthermore, we note that there is an equivalence between $\text{MB-MSPBE}$ and MSPBE. That is, if a best non-linear model is learned from the data generated from some policy $b$ and $\zeta$ in the search control process equals $b$'s stationary feature vector distribution $\mu$, then $\text{MB-MSPBE}$ is the MSPBE.

\begin{proposition}\label{proposition_62}
If
\begin{enumerate}
\item Assumption \ref{behavior policy assumption} and \ref{search control assumption} hold.
\item $\zeta = \mu$
\item $\hat\bx = \hat\bx^*$ and $\hat r = \hat r^*$
\end{enumerate}
Then $\text{MB-MSPBE}(\bw) = \text{MSPBE}(\bw)$.
\end{proposition}
The above proposition does not hold for the best linear model for the same reason elaborated in proposition \ref{TD(0) planning}.

% Note that $\bw^*$ is the same as the TD(0) planning solution if $\bd(\bphi, a) = \bF_a \bphi, \thinspace \thinspace \hat r(\bphi, a) = \bb_a^\top \bphi$, but since Gradient Dyna optimizes this objective by gradient descent, the numerical radius condition is no longer required for the convergence of the algorithm.

\begin{figure}\centering
  \includegraphics[width=7.5cm]{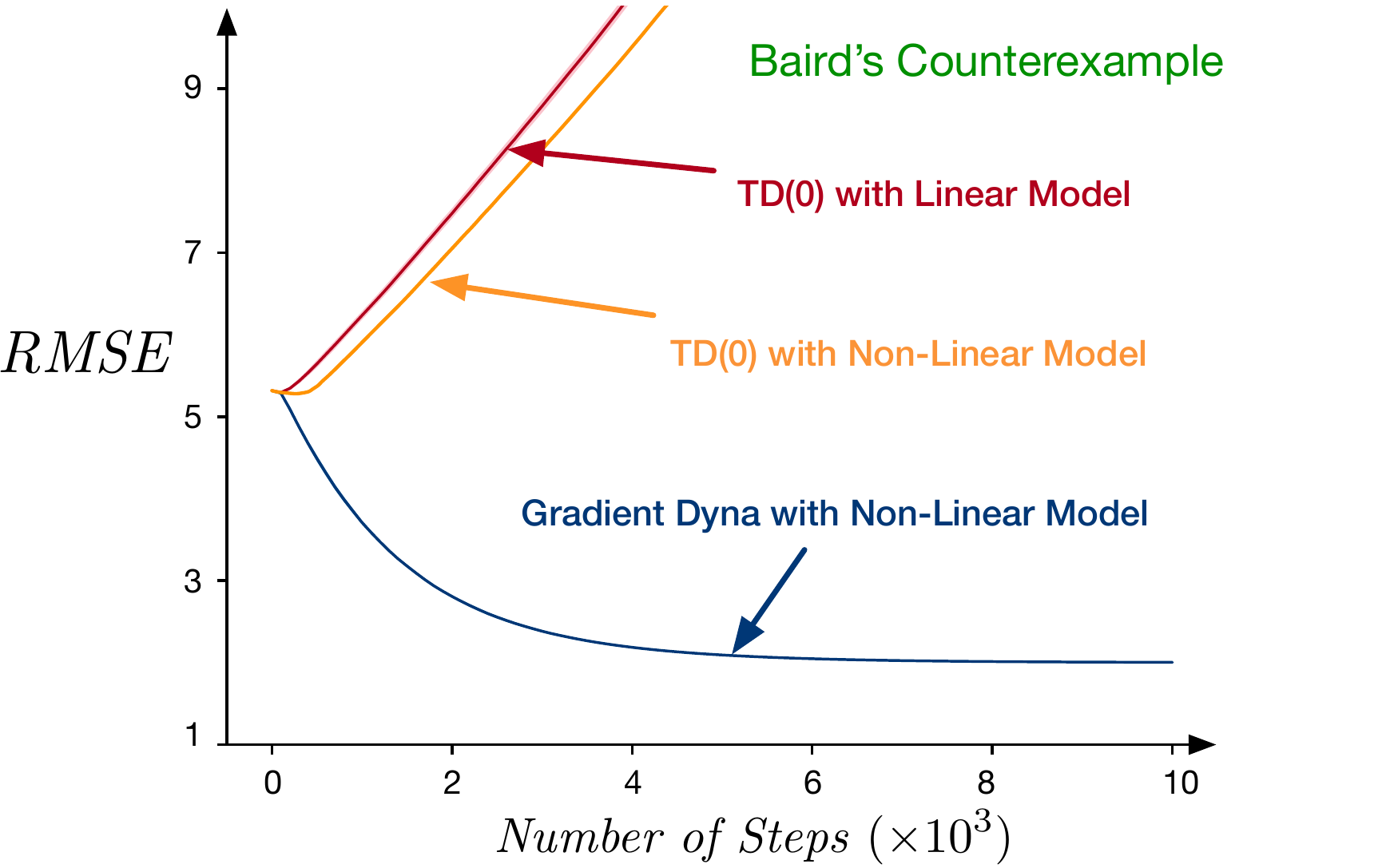}
  \caption{Gradient Dyna vs TD(0) in \textit{Baird's counterexample}: Gradient Dyna remains stable, whereas TD(0) algorithm diverges. The reported curve is the average of 30 runs and the standard deviation is too small to be visible clearly}
  \label{fig:BCE}
\end{figure}

Let's now consider algorithms that can be used to minimize this objective. Consider the gradient of MB-MSPBE: $
\nabla \text{MB-MSPBE}(\bw)
 = \mathbb{E}_\zeta[(\gamma \hat\bx(\bphi_k, A_k) - \bphi_k) \bphi_k^\top] \mathbb{E}_\zeta[\bphi_k \bphi_k^\top]^{-1} \bE_\zeta [\Delta_k \bphi_k] 
$. We have a product of three expectations and, therefore, we cannot simply use one sample to obtain an unbiased estimate of the gradient. In order to obtain an unbiased estimate of the gradient, we could either draw three independent samples or learn the product of the the last two factors using a linear least-square method. GTD methods take the second route leading to an algorithm with $O(m)$ complexity in which two sets of parameters are mutually related in their updates. However, if one uses a linear model, the computational complexity for storing and using the model is already $O(m^2)$. For a non-linear model, depending on the parameterization choices, the complexity can be either smaller or greater than $O(m^2)$. Thus, a planning algorithms with $O(m^2)$ complexity can be an acceptable choice. This leads to two choices: we can sample the three expectations and and then multiply them to produce an unbiased estimate of the gradient. Note that this would still lead to an $O(m^2)$ algorithm as the matrix inversion can be done in $O(m^2)$ using Sherman-Morrison formula. Alternatively, we can use the linear least-square method to estimate the first two expectations and sample the third one. Compared with GTD methods, there are still two sets of parameters but their updates are not mutually dependent, which potentially leads to faster convergence. Although both of these approaches have $O(m^2)$ complexity, we adopt the second approach, which is summarized in algorithm \ref{GDP}. We now present the convergence theorem for the proposed algorithm, which is followed by its empirical evaluation. The reader can refer to the supplementary materials for the proof of the theorem.

\begin{algorithm}[tb] 
\caption{Gradient Dyna Algorithm}
\label{GDP}
\textbf{Input}: $\bw_0, \bV_0$, policy $\pi$, feature vector distribution $\zeta$, expectation model $\{\hat\bx, \hat r\}$, stepsizes $\alpha_k, \beta_k$ for $k = 1, 2, \cdots$\\
\textbf{Output}: $\bw_k$
\begin{algorithmic}[1] %[1] enables line numbers
\FOR {$k = 1, 2, \cdots$}
\STATE Sample $\bphi_k \sim \zeta (\cdot)$
\STATE Sample $A_k \sim \pi(\cdot | \bphi_k)$
% \STATE ($\bV_k$ approximates the product of the first two expectations in the gradient of MB-MSPBE.)
\STATE $\bw_{k+1} \leftarrow \bw_k - \alpha_k \bV_{k} \Delta_k  \bphi_k$
\STATE $\bV_{k+1} \leftarrow \bV_{k} + \beta_k ((\gamma \hat\bx(\bphi_k, A_k)  - \bphi_k) \bphi_k^\top - \bV_k \bphi_k \bphi_k^\top )$
%     \begin{align*}
%     \bw_{k+1} & \leftarrow \bw_k - \alpha_k \bV_{k} \Delta_k  \bphi_k\\
%     \bV_{k+1} & \leftarrow \bV_{k} + \beta_k ((\gamma \hat\bx(\bphi_k, A_k)  - \bphi_k) \bphi_k^\top - \\
%     & \bV_k \bphi_k \bphi_k^\top )
% \end{align*} 
\ENDFOR
\end{algorithmic}
\end{algorithm}

\begin{figure*}\centering
% \vspace{-1.5cm}
  \includegraphics[width=14cm]{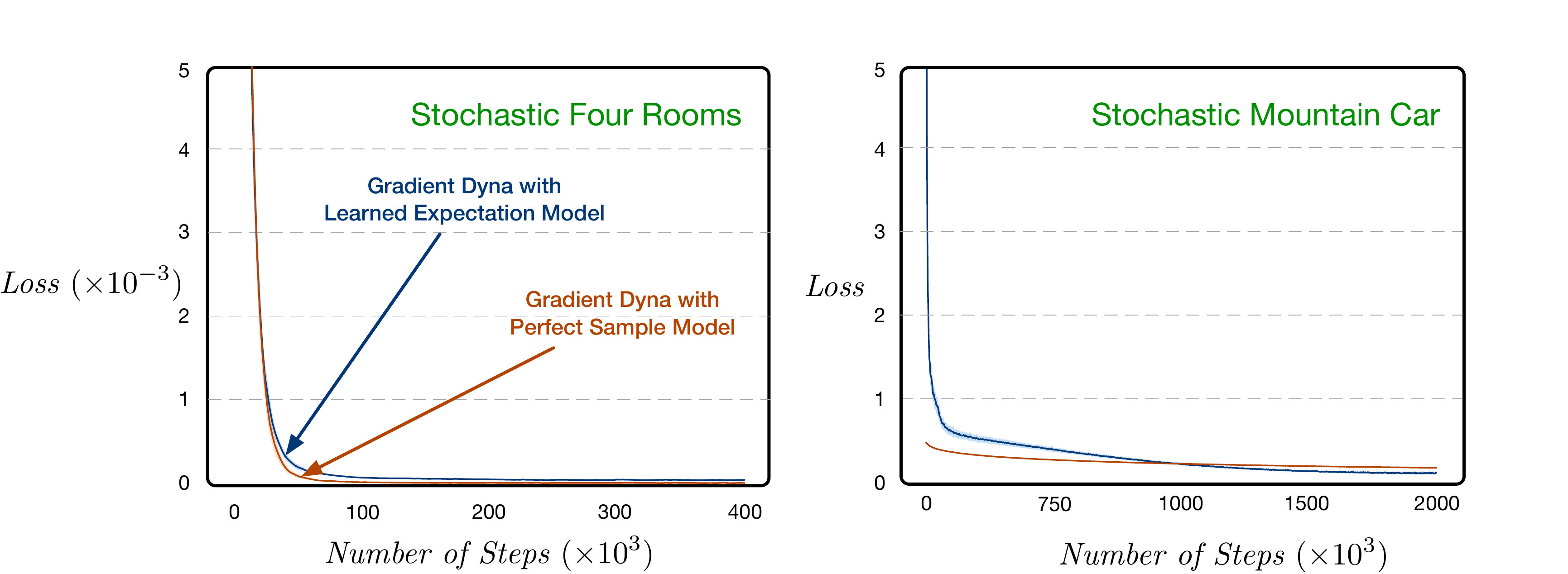}
  \caption{Gradient Dyna with a learned expectation model remains stable and converges to the off-policy LSTD solution. The reported curve is the average of 30 runs and the standard deviation is too small to be visible clearly}
  \label{fig:MC}
\end{figure*}

\begin{theorem}[Convergence of Gradient Dyna Algorithm]\label{Convergence of GTD-planning-v1}

Consider algorithm \ref{GDP}.
If
\begin{enumerate}
    \item Assumptions \ref{bounded feature vector}, \ref{search control assumption} and \ref{MB-MSPBE assumption} hold
    \item $\alpha_k > 0, \beta_k > 0$, $\sum_{k=0}^\infty \alpha_k = \infty, \sum_{k=0}^\infty \beta_k = \infty$, $\sum_{k=0}^\infty \alpha_k^2 < \infty, \sum_{k=0}^\infty \beta_k^2 < \infty$
\end{enumerate}
Then for any initial weight vector $\bw_0$, $\lim_{k \rightarrow \infty} \bw_k = \bw^*$ w.p. 1.
\end{theorem}

\section{Experiments}

The goal of the experiment section is to validate the theoretical results and investigate how Gradient Dyna algorithm performs in practice. Concretely, we seek to answer the following questions: 1) is the proposed planning algorithm stable for the non-linear model choice, especially when the model is learned online and 2) what solution does the proposed planning algorithm converge to.

\subsection{Divergence in TD(0)}
Our first experiment is designed to illustrate the divergence issue with TD(0) update. We use \textit{Baird's counterexample} \cite{baird1995residual,sutton2018reinforcement} - a classic example to highlight the off-policy divergence problem with the model-free TD(0) algorithm. The policy $b$ used to learn the model is arranged to be the same as the behavior policy in the counterexample, whereas the policy $\pi$ to be evaluated is arranged to be the same as the counterexample's target policy. For TD(0) with linear model, we initialize the matrix $\bF_a$ and vector $\bb_a$ for all $a$ to be zero. 
For TD(0) with non-linear model and Gradient Dyna, we use a neural network with one hidden layer of 200 units as the non-linear model. We initialize the non-linear model using Xavier initialization \cite{glorot2010understanding}.
The parameter $\bw$ for the estimated value function is initialized as proposed in the counterexample. The model is learned in an online fashion, that is, we use only the most recent sample to perform a gradient-descent update on the mean-square error. The search-control process is also restricted to generate the last-seen feature vector, which is then used with an $a \sim \pi$ to simulate the next feature vector. The resulting simulated transition is used to apply the planning update. The evaluation metric is the Root Mean Square Error (RMSE): $\sqrt{\sum_{s} (\hat{v}(\bx(s), \bw) - v_\pi (s))^2 / |\States|}$. The results are reported for hyperparameters chosen based on RMSE over the latter half of a run. In Figure \ref{fig:BCE}, we see that TD(0) updates with both linear and non-linear expectation model cause the value function to diverge. In contrast, Gradient Dyna remains sound and converges to the RMSE of 2.0. % Interestingly, stable model-free methods also converge to the same RMSE value (not shown here) \cite{sutton2018reinforcement}.

%   For this and all experiments in this paper, only one planning step per time step is used. We sweep hyper-parameters over XXX. For GDP we tried both linear and non-linear model. Similarly, model is learned online. We sweep hyper-parameters XXX.

\subsection{Convergence in Practice}
In this set of experiments, we want to investigate how Gradient Dyna algorithm performs in practice. We evaluate the proposed method for the non-linear model choice in two simple yet illustrative domains: stochastic variants of Four Rooms \cite{sutton1999between,DBLP:journals/corr/abs-1811-02597} and Mountain Car \cite{sutton1996generalization}. Similar to the previous experiment, the model is learned online. Search control, however, samples uniformly from the recently visited 1000 feature vectors to approximate the i.i.d. assumption in Theorem \ref{Convergence of GTD-planning-v1}. 

We modified the Four Rooms domain by changing the states on the four corners to terminal states. The reward is zero everywhere except when the agent transitions into a terminal state, where the reward is one. The episode starts in one of the non-terminal states uniform randomly. The policy $b$ to generate the data for learning the model takes all actions with equal probability, whereas the policy $\pi$ to be evaluated constitutes the shortest path to the top-left terminal state and is deterministic. We used tile coding \cite{sutton1996generalization} to obtain feature vector($4$ $2\times2$ tilings). In mountain car, the policy $b$ used to generate the data is the standard energy-pumping policy with 50$\%$ randomness \cite{le2017learning}, where the policy $\pi$ to be evaluated is also the standard energy-pumping policy but with no randomness. We again used tile coding to obtain feature vector ($8$ $8\times8$ tilings). We inject stochasticity in the environments by only executing the chosen action $70\%$ of the times, whereas a random action is executed 30$\%$ of the time. In both experiments, we do one planning step for each sample collected by the policy $b$. As noted in proposition 6.1, if we have $\zeta = \mu$, the minimizer of MB-MSPBE for the best non-linear model is the off-policy LSTD solution $\bA_{\text{LSTD}}^{-1} \bc_{\text{LSTD}}$, $\bA_{\text{LSTD}} = \bE_b[\rho_t \bx_t (\bx_t - \gamma \bx_{t+1})^{\top}]$, $\bc_{\text{LSTD}} = \bE_b[\rho_t R_{t+1} \bx_t]$ \cite{yu2010convergence}. Therefore, for both domains, we run the off-policy LSTD algorithm for 2 million time-steps and use the resulting solution as the evaluation metric: $Loss = \Vert \bA_{\text{LSTD}} \bw - \bc_{\text{LSTD}}\Vert_2^2$. 

The results are reported for hyper-parameters chosen according to the LSTD-solution based loss over the latter half of a run. In Figure \ref{fig:MC}, we see that Gradient Dyna remains stable and converges to off-policy LSTD solution in both domains. % We also report the performance of Gradient Dyna with a perfect sample model.

%\section{Expectation-based planning using learned features}
%Here we will talk about how linearity is not a constraint but a design decision and how feature learning is moved to state update function. We will also give a sketch of the \textit{two-part approximation} where representation is detached from value-function and how model/planning live in the state-feature space.

\section{Conclusion}
In this paper, we proposed a sound way of using the expectation models for planning and showed that it is equivalent to planning with distribution models if the state value function is linear in feature vector. We made a theoretical argument for non-linear expectation models to be the parametrization of choice even if the value-function is linear. Lastly, we proposed Gradient Dyna, a model-based policy evaluation algorithm with convergence guarantees, and empirically demonstrated its effectiveness.

\section*{Acknowledgements}
We would like to thank Huizhen Yu, Sina Ghiassian, Banafsheh Rafiee and Khurram Javed for useful discussions and feedbacks.

% \section{\LaTeX{} and Word Style Files}\label{stylefiles}

% The \LaTeX{} and Word style files are available on the IJCAI--19
% website, \url{http://www.ijcai19.org}.
% These style files implement the formatting instructions in this
% document.

% The \LaTeX{} files are {\tt ijcai19.sty} and {\tt ijcai19.tex}, and
% the Bib\TeX{} files are {\tt named.bst} and {\tt ijcai19.bib}. The
% \LaTeX{} style file is for version 2e of \LaTeX{}, and the Bib\TeX{}
% style file is for version 0.99c of Bib\TeX{} ({\em not} version
% 0.98i). The {\tt ijcai19.sty} style differs from the {\tt
% ijcai18.sty} file used for IJCAI--18.

% The Microsoft Word style file consists of a single file, {\tt
% ijcai19.doc}. This template differs from the one used for
% IJCAI--18.

% These Microsoft Word and \LaTeX{} files contain the source of the
% present document and may serve as a formatting sample.  

% Further information on using these styles for the preparation of
% papers for IJCAI--19 can be obtained by contacting {\tt
% pcchair@ijcai19.org}.


\newpage

\bibliographystyle{named}
\bibliography{ijcai19}
\thispagestyle{empty}

% \mbox{}



\section*{Supplementary material}

# Supplementary Materials

## 1 Proofs

**Proof of Proposition 5.1**
For real environment the fixed point is

$$\mathbf{w}_{\text{real}} = \mathbb{E}_b[\rho_t \mathbf{x}_t (\mathbf{x}_t - \gamma \mathbf{x}_{t+1})^\top]^{-1} \mathbb{E}_b[\rho_t R_{t+1} \mathbf{x}_t]$$

For best linear model we showed in the paper the solution is

$$\mathbf{w}_{\text{linear}} = (\mathbf{I} - \gamma \mathbf{F}^\top)^{-1} \mathbf{b}$$

where $\mathbf{F} = \mathbb{E}_\zeta[\mathbf{F}^*_{A_k} \boldsymbol{\phi}_k \boldsymbol{\phi}_k^\top] \mathbb{E}_\zeta[\boldsymbol{\phi}_k \boldsymbol{\phi}_k^\top]^{-1}, \mathbf{b} = \mathbb{E}_\zeta[\boldsymbol{\phi}_k \boldsymbol{\phi}_k^\top]^{-1} \mathbb{E}_\zeta[\boldsymbol{\phi}_k \boldsymbol{\phi}_k^\top \mathbf{b}^*_{A_k}]$.
Plug in $\mathbf{F}, \mathbf{b}$, we have

$$\mathbf{w}_{\text{linear}} = \mathbb{E}_\zeta[\boldsymbol{\phi}_k \boldsymbol{\phi}_k^\top (\mathbf{I} - \gamma \mathbf{F}^{*\top}_{A_k})]^{-1} \mathbb{E}_\zeta[\boldsymbol{\phi}_k \boldsymbol{\phi}_k^\top \mathbf{b}^*_{A_k}]$$

where

$$\mathbb{E}_\zeta[\boldsymbol{\phi}_k \boldsymbol{\phi}_k^\top (\mathbf{I} - \gamma \mathbf{F}^{*\top}_{A_k})]$$
$$= \sum_a (\sum_{\boldsymbol{\phi}} \mu(\boldsymbol{\phi}) \boldsymbol{\phi} \boldsymbol{\phi}^\top \pi(a|\boldsymbol{\phi}))(\mathbf{I} - \gamma \mathbb{E}_b[\mathbb{I}(A_t = a) \mathbf{x}_t \mathbf{x}_t^\top]^{-1} \mathbb{E}_b[\mathbb{I}(A_t = a) \mathbf{x}_t \mathbf{x}_{t+1}^\top])$$
$$= \sum_a \mathbb{E}_b[\mathbb{I}(A_t = a) \rho_t \mathbf{x}_t \mathbf{x}_t^\top] \mathbb{E}_b[\mathbb{I}(A_t = a) \mathbf{x}_t \mathbf{x}_t^\top]^{-1} \mathbb{E}_b[\mathbb{I}(A_t = a) \mathbf{x}_t (\mathbf{x}_t - \gamma \mathbf{x}_{t+1})^\top]$$

and

$$\mathbb{E}[\boldsymbol{\phi}_k \boldsymbol{\phi}_k^\top \mathbf{b}^*_{A_k}]$$
$$= \sum_a (\sum_{\boldsymbol{\phi}} \mu(\boldsymbol{\phi}) \boldsymbol{\phi} \boldsymbol{\phi}^\top \pi(a|\boldsymbol{\phi})) \mathbb{E}_b[\mathbb{I}(A_t = a) \mathbf{x}_t \mathbf{x}_t^\top]^{-1} \mathbb{E}_b[\mathbb{I}(A_t = a) \mathbf{x}_t R_{t+1}]$$
$$= \sum_a \mathbb{E}_b[\mathbb{I}(A_t = a) \rho_t \mathbf{x}_t \mathbf{x}_t^\top] \mathbb{E}_b[\mathbb{I}(A_t = a) \mathbf{x}_t \mathbf{x}_t^\top]^{-1} \mathbb{E}_b[\mathbb{I}(A_t = a) \mathbf{x}_t R_{t+1}]$$

We see that $\mathbf{w}_{\text{linear}} \neq \mathbf{w}_{\text{real}}$ in general.

For the best non-linear model, the fixed-point is

$$\begin{aligned}
\mathbf{w}_{\text{non-linear}} &= \mathbb{E}_\zeta[\boldsymbol{\phi}_k(\boldsymbol{\phi}_k - \gamma\hat{\mathbf{x}}^*(\boldsymbol{\phi}_k, A_k))^\top]^{-1}\mathbb{E}_\zeta[\hat{r}^*(\boldsymbol{\phi}_k, A_k)\boldsymbol{\phi}_k] \\
&= (\sum_{\boldsymbol{\phi}} \mu(\boldsymbol{\phi})\boldsymbol{\phi}\sum_a \pi(a|\boldsymbol{\phi})(\boldsymbol{\phi} - \gamma\mathbb{E}_\zeta[\boldsymbol{\phi}'|\boldsymbol{\phi}, a])^\top)^{-1} \\
&\quad (\sum_{\boldsymbol{\phi}} \mu(\boldsymbol{\phi})\boldsymbol{\phi}\sum_a \pi(a|\boldsymbol{\phi})\mathbb{E}_\zeta[R'|\boldsymbol{\phi}, a]) \\
&= \mathbb{E}_b[\rho_t\mathbf{x}_t(\mathbf{x}_t - \gamma\mathbf{x}_{t+1})^\top]^{-1}\mathbb{E}_b[\rho_t R_{t+1}\mathbf{x}_t] \\
&= \mathbf{w}_{\text{real}}
\end{aligned}$$

**Q.E.D.**

**Proof of Proposition 6.1**
From what we have shown in the above proposition we know $\mathbb{E}_\zeta[\Delta_k\boldsymbol{\phi}_k] = \mathbb{E}_b[\rho_t\delta_t\mathbf{x}_t]$, immediately

$$\begin{aligned}
\text{MB-MSPBE}(\mathbf{w}) &= \mathbb{E}[\Delta_k\boldsymbol{\phi}_k]^\top\mathbb{E}[\boldsymbol{\phi}_k\boldsymbol{\phi}_k^\top]^{-1}\mathbb{E}[\Delta_k\boldsymbol{\phi}_k] \\
&= \mathbb{E}_b[\rho_t\delta_t\mathbf{x}_t]^\top\mathbb{E}_b[\mathbf{x}_t\mathbf{x}_t^\top]^{-1}\mathbb{E}_b[\rho_t\delta_t\mathbf{x}_t] \\
&= \text{MSPBE}(\mathbf{w})
\end{aligned}$$

**Q.E.D.**

**Proof of Theorem 6.1**
Note that two updates are unilateral. The second update is a standard SGD update, which doesn't rely on the first one and thus can be shown to converge under our assumptions using either o.d.e approach or proposition 4.1 of [Bertsekas and Tsitsiklis, 1996]. Convergence proof for this update thus is omitted.
The convergent point for $\mathbf{V}_k$ is

$$\mathbf{V} \doteq \lim_{k \to \infty} \mathbf{V}_k = \mathbb{E}_\zeta[(\gamma\hat{\mathbf{x}}(\boldsymbol{\phi}_k, A_k) - \boldsymbol{\phi}_k)\boldsymbol{\phi}_k^\top]\mathbb{E}_\zeta[\boldsymbol{\phi}_k\boldsymbol{\phi}_k^\top]^{-1} = -\mathbf{A}^\top\mathbf{C}^{-1}$$

For the convergence of the first update, we first apply the last extension in section 2.2 of [Borkar, 2008] which states the diminishing bounded noise won't change convergence of the update equation. For this purpose write the first update as three terms, the expectation of the update in the limit, the martingale and the additional error term due to the inaccuracy from $\mathbf{V}_k$.
The first update can be written as

$$\mathbf{w}_{k+1} = \mathbf{w}_k + \alpha_k(-\mathbf{V}\mathbb{E}_\zeta[\Delta_k\boldsymbol{\phi}_k] + \mathbf{V}(\mathbb{E}_\zeta[\Delta_k\boldsymbol{\phi}_k] - \Delta_k\boldsymbol{\phi}_k) + (\mathbf{V} - \mathbf{V}_k)\Delta_k\boldsymbol{\phi}_k)$$

Let $h(\mathbf{w}) = -\mathbf{V}\mathbb{E}_\zeta[\Delta_k\boldsymbol{\phi}_k]$, $M_{k+1} = \mathbf{V}(\Delta_k\boldsymbol{\phi}_k - \mathbb{E}_\zeta[\Delta_k\boldsymbol{\phi}_k])$ and $\epsilon_k = (\mathbf{V}_k - \mathbf{V})\Delta_k\boldsymbol{\phi}_k$. Therefore

$$\mathbf{w}_{k+1} = \mathbf{w}_k + \alpha_k(h(\mathbf{w}_k) + M_{k+1} + \epsilon_k) \tag{1}$$

We now apply corollary 3 in section 2 of [Borkar, 2008] to show the convergence and the convergent point. Let's now define a function $V(\mathbf{w}) \doteq \text{MB-MSPBE}(\mathbf{w}) - \min_{\mathbf{w}} \text{MB-MSPBE}(\mathbf{w})$. Note that 1) $V : \mathbb{R}^d \mapsto [0, \infty)$ is continuously differentiable 2) $\lim_{\|\mathbf{w}\| \to \infty} V(\mathbf{w}) = \infty$, and 3) $H \doteq \{\mathbf{w} \in \mathbb{R}^d : V(\mathbf{w}) = 0\} = \{\arg\min_{\bar{\mathbf{w}}} \text{MB-MSPBE}(\bar{\mathbf{w}})\}$ because minimizing $J$ only has one solution, 4) it's true that $<h(\mathbf{w}), \nabla V(\mathbf{w})> = -\|h(\mathbf{w})\|^2 \leq 0$ with the inequality iff $\mathbf{w} \in H$. Corollary 3 concludes that $\{\mathbf{w}_k\}$ a.s. converge to an internally chain transitive invariant set contained in $H$, which in our case is a singleton $\{\arg\min_{\mathbf{w}} \text{MB-MSPBE}(\mathbf{w})\}$, if following conditions are true:

1. The map $h : \mathbb{R}^d \mapsto \mathbb{R}^d$ is Lipschitz: $\|h(\mathbf{x}) - h(\mathbf{y})\| \leq L\|\mathbf{x} - \mathbf{y}\|$ for some $0 < L < 1$.

2. Stepsizes $\{\alpha_k\}$ are positive scalars satisfying $\sum_k \alpha_k = \infty, \sum_k \alpha_k^2 < \infty$

3. $\{M_k\}$ is a martingale difference sequence with respect to the increasing family of $\sigma$-fields

$$\mathcal{F}_k \doteq \sigma(\mathbf{w}_j, M_j, j \leq k) \quad (2)$$

    That is, $\mathbb{E}[M_{k+1}|\mathcal{F}_k] = 0$ a.s., $k \geq 0$. Furthermore, $\{M_k\}$ are square-integratable with $\mathbb{E}[\|M_k\|^2|\mathcal{F}_k] \leq K(1 + \|\mathbf{w}_k\|^2)$ for some $K > 0$,

4. The iterates of (1) remain bounded a.s.,

5. $\{\epsilon_k\}$ is a deterministic or random bounded sequence which is $o(1)$.

Let's now verify these conditions
1: $h$ is clearly Lipschitz with $L = \|\mathbf{VA}\|$.
2: this condition is satisfied in the assumption 2 of theorem 6.1
3: $\mathbb{E}[M_{k+1}|\mathcal{F}_k] = \mathbf{V}\mathbb{E}_\zeta[\Delta_k \boldsymbol{\phi}_k - \mathbb{E}_\zeta[\Delta_k \boldsymbol{\phi}_k]|\mathbf{w}_k] = 0$. The square-integratability of $\{M_k\}$ can be shown by applying triangle inequality and boundedness of $\boldsymbol{\phi}_k$ and the model $\{\hat{\mathbf{x}}, \hat{r}\}$ (the model ouput is bounded as the co-domain of the $\hat{\mathbf{x}}$ and $\hat{r}$ are $\mathbb{R}^m$ and $\mathbb{R}$ and the domain of $\hat{\mathbf{x}}$ and $\hat{r}$ is a finite set as we assumed finite state finite action MDP)
4: This can be satisfied by applying theorem 7 in section 3 of [Borkar, 2008] which requires us to check one additional condition: the functions $h_c(\mathbf{z}) \doteq h_c(c\mathbf{z})/c, c \geq 1, \mathbf{z} \in \mathbb{R}^d$ satisfy $h_c(\mathbf{z}) \to h_\infty(\mathbf{z})$ as $c \to \infty$, uniformly on compacts for some $h_\infty \in C(\mathbb{R}^d)$. Furthermore, the o.d.e. $\mathbf{z}(t) = h_\infty(\mathbf{z}_t)$ has the origin as its unique globally asymptotically stable equilibrium. This is true as $h_\infty(\mathbf{x}) = \mathbf{VAx}$. And $\mathbf{VA} = -\mathbf{A}^\top \mathbf{C}^{-1}\mathbf{A}$ is negative definite as both $\mathbf{A}$ and $\mathbf{C}$ are non-singular and $\mathbf{C}^{-1}$ is positive definite. Thus $\mathbf{z}(t) = h_\infty(\mathbf{z}_t)$ has the origin as its unique globally asymptotically stable equilibrium.
5: since $\lim_{k \to \infty} \mathbf{V}_k = \mathbf{V}$ a.s. and $\Delta_k$ and $\boldsymbol{\phi}_k$ are bounded because of the boundedness of $\mathbf{w}_k$ and our assumption on the boundedness of $\boldsymbol{\phi}_k$ and the model, $\lim_{k \to \infty}(\mathbf{V}_k - \mathbf{V})\Delta_k \boldsymbol{\phi}_k = 0$ a.s. **Q.E.D.**

## 2 Hyperparameter Settings

Here we present the details of the experiments. For each algorithm, we swept over a range of hyper-parameters (specified below) with 10 runs for each setting. Results are reported for hyperparameters chosen based on the specified evaluation metric over the latter half of a run.

**Baird's Counterexample (section 7.1)**
    Hyper-parameters ranges for TD(0)-based planning
        $\alpha \in \{0.03, 0.01, 0.003, 0.001, 0.0003, 0.0001\}$
        model-learning rate $\in \{0.1, 0.01\}$
    Hyper-parameters ranges for GDP
        $\alpha \in \{0.1, 0.01, 0.001\}$
        $\beta \in \{0.1, 0.01, 0.001\}$
        model-learning rate $\in \{0.1, 0.01\}$

**Stochastic Mountain Car & Four Room Domain (section 7.2)**
    Hyper-parameters ranges for GDP
        $\alpha \in \{0.1, 0.01, 0.001\}$
        $\beta \in \{0.1, 0.01, 0.001\}$
        model-learning rate $\in \{0.1, 0.01, 0.001\}$



\end{document}